\documentclass[10pt,twocolumn,letterpaper]{article}

\usepackage[pagenumbers]{cvpr} 

\usepackage{graphicx}
\usepackage{amsmath}
\usepackage{amssymb}
\usepackage{enumitem}
\usepackage{booktabs}
\usepackage{bm}

\usepackage{algorithm}
\usepackage{algorithmic}
\usepackage{multirow}
\usepackage{mathrsfs}

\usepackage[pagebackref,breaklinks,colorlinks]{hyperref}

\usepackage[capitalize]{cleveref}
\crefname{section}{Sec.}{Secs.}
\Crefname{section}{Section}{Sections}
\Crefname{table}{Table}{Tables}
\crefname{table}{Tab.}{Tabs.}

\def\cvprPaperID{1002} 
\def\confName{CVPR}
\def\confYear{2022}

\begin{document}

\title{Generalizable Cross-modality Medical Image Segmentation via Style Augmentation and Dual Normalization}


\author{
    Ziqi Zhou\textsuperscript{\rm 1}\quad
    Lei Qi\textsuperscript{\rm 2}\quad
    Xin Yang\textsuperscript{\rm 3}\quad
    Dong Ni\textsuperscript{\rm 3}\quad
    Yinghuan Shi\textsuperscript{\rm 1}\thanks{Corresponding author: Yinghuan Shi.~This work was supported by National Key Research and Development Program of China (2019YFC0118300), NSFC Major Program (62192783), CAAI-Huawei MindSpore Project (CAAIXSJLJJ-2021-042A), China Postdoctoral Science Foundation Project (2021M690609), and Jiangsu Natural Science Foundation Project (BK20210224).}~~\thanks{Ziqi Zhou and Yinghuan Shi are with the State Key Laboratory for Novel Software Technology and National Institute of Healthcare Data Science, Nanjing University, China. Lei Qi is with the School of Computer Science and Engineering, Southeast University, China. Xin Yang and Dong Ni are with the National-Regional Key Technology Engineering Laboratory for Medical Ultrasound, School of Biomedical Engineering, Health Science Center, the Medical Ultrasound Image Computing Lab, the Marshall Laboratory of Biomedical Engineering, Shenzhen University, China.}\\
    \textsuperscript{\rm 1}Nanjing University\quad\textsuperscript{\rm 2}Southeast University\quad\textsuperscript{\rm 3}Shenzhen University\\
    \vspace{0.1cm}
    {\tt\small zhouzq@smail.nju.edu.cn}\quad{\tt\small qilei@seu.edu.cn}\quad{\tt\small \{xinyang, nidong\}@szu.edu.cn}\quad{\tt\small syh@nju.edu.cn}
}
\maketitle

\begin{abstract}
  For medical image segmentation, imagine if a model was only trained using MR images in source domain, how about its performance to directly segment CT images in target domain? This setting, namely generalizable cross-modality segmentation, owning its clinical potential, is much more challenging than other related settings, e.g., domain adaptation. To achieve this goal, we in this paper propose a novel dual-normalization model by leveraging the augmented source-similar and source-dissimilar images during our generalizable segmentation. To be specific, given a single source domain, aiming to simulate the possible appearance change in unseen target domains, we first utilize a nonlinear transformation to augment source-similar and source-dissimilar images. Then, to sufficiently exploit these two types of augmentations, our proposed dual-normalization based model employs a shared backbone yet independent batch normalization layer for separate normalization. Afterward, we put forward a style-based selection scheme to automatically choose the appropriate path in the test stage. Extensive experiments on three publicly available datasets, i.e., BraTS, Cross-Modality Cardiac, and Abdominal Multi-Organ datasets, have demonstrated that our method outperforms other state-of-the-art domain generalization methods. Code is available at \url{https://github.com/zzzqzhou/Dual-Normalization}.
\end{abstract}

\section{Introduction}
\begin{figure}[t]
    \centering
    \begin{subfigure}{0.49\linewidth}
        \centering
        \includegraphics[width=1.0\columnwidth]{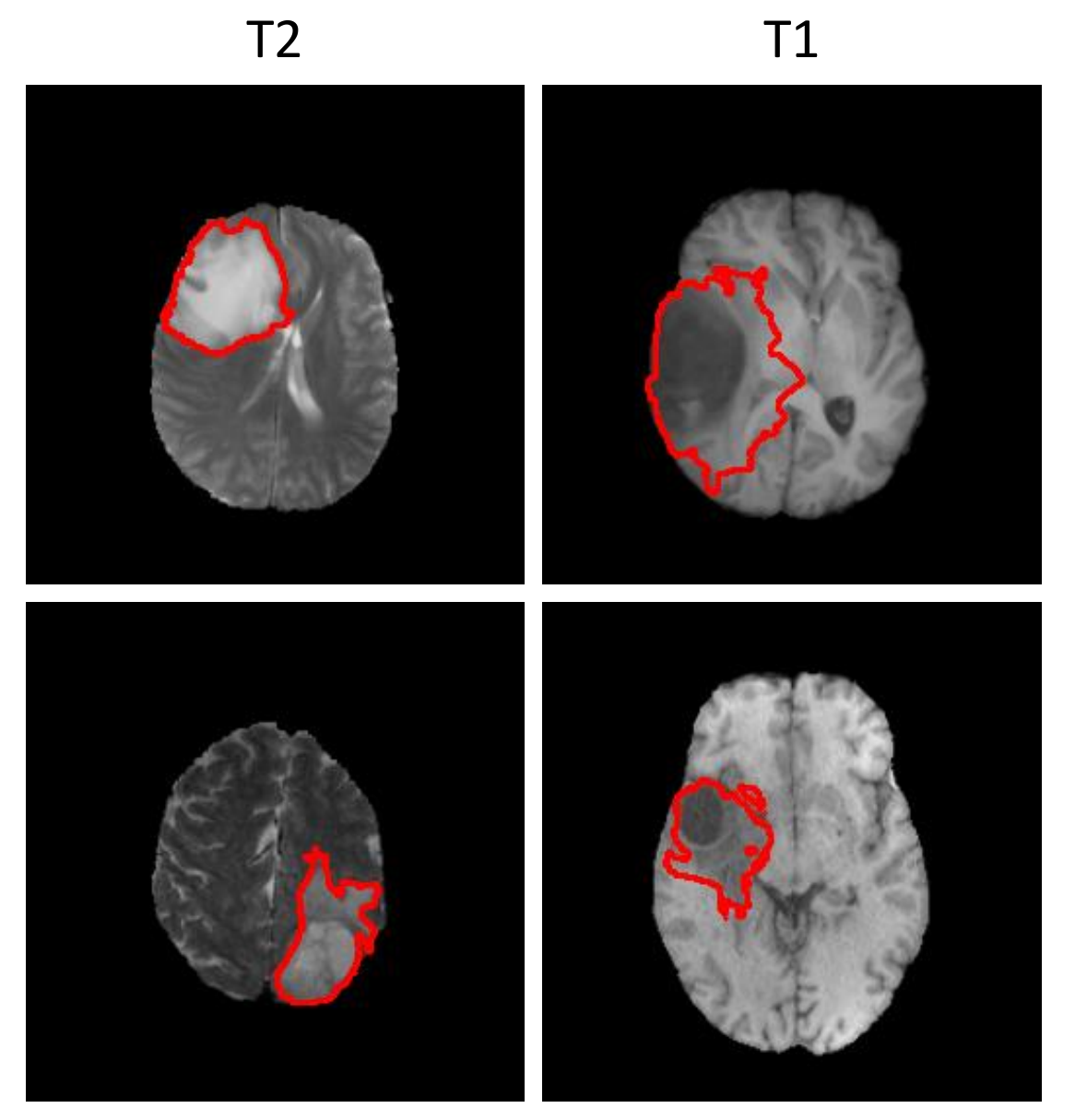}
        \caption{Example slices of T2 \& T1.}
        \label{fig1_a}
    \end{subfigure}
    \begin{subfigure}{0.49\linewidth}
        \centering
        \includegraphics[width=1.05\columnwidth]{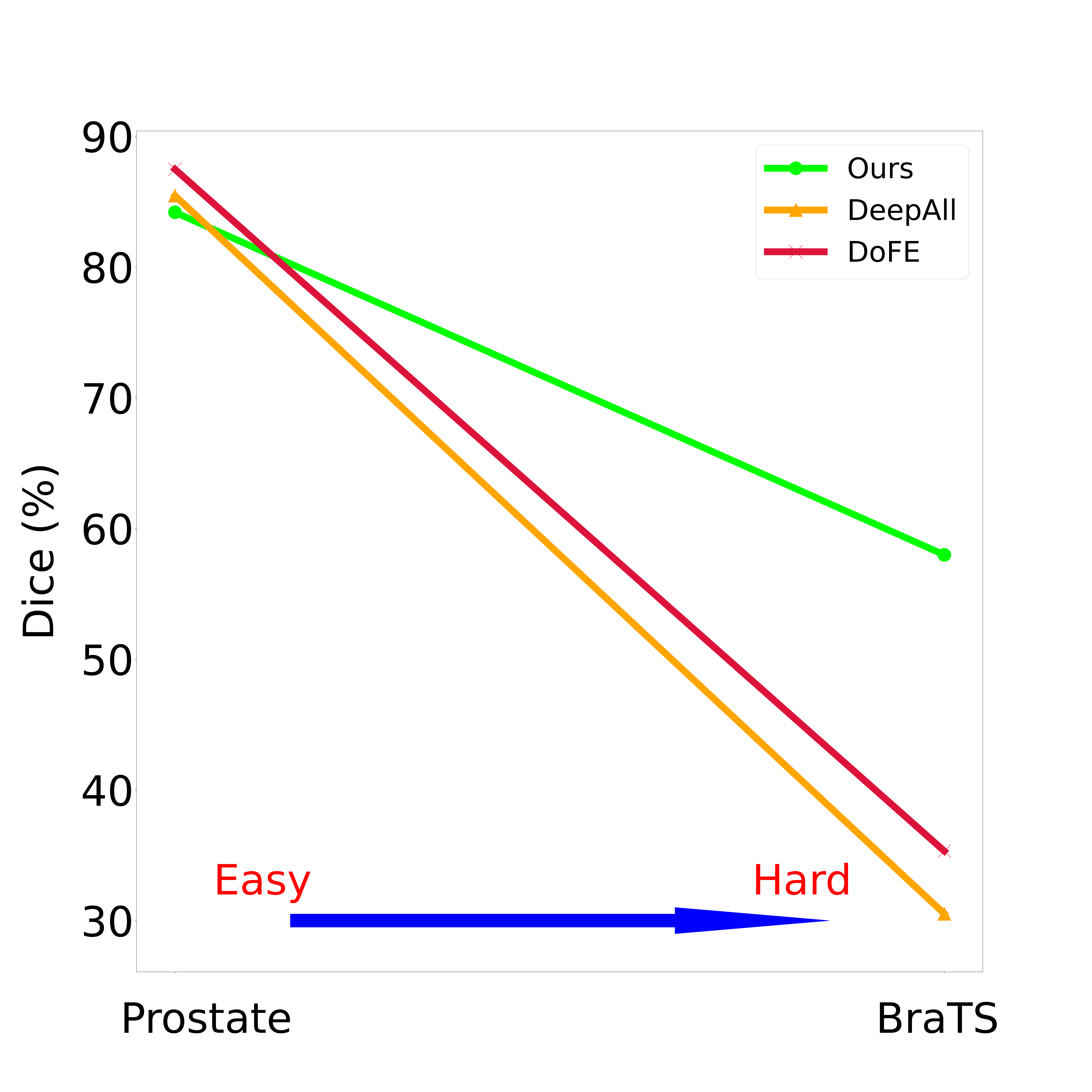}
        \caption{Performance.}
        \label{fig1_b}
    \end{subfigure}
    \caption{(a) Example slices from BraTS dataset; (b) Comparison of our method with ``DeepAll'' and ``DoFE'' methods on the \textbf{Cross-center} prostate segmentation task and the \textbf{Cross-modality} brain tumor segmentation task.}
    \label{fig1}
    \vspace{-10pt}
\end{figure}
In recent years, profound progress in medical image segmentation has been achieved by deep convolutional neural networks~\cite{ronneberger2015u,milletari2016v,li2018h}. Benefiting from these recent efforts, the accuracy of segmentation on medical images has now been substantially improved. Despite their success, the distribution shift between training (or labeled) and test (or unlabeled) data usually results in a severe performance degeneration during the deployment of trained segmentation models. The reason for distribution shift typically come from different aspects, \eg, different acquisition parameters, various imaging methods or diverse modalities.

To fight against domain shift, several practical settings have been investigated, among which unsupervised domain adaptation (UDA) based segmentation~\cite{hoffman2018cycada,chen2019crdoco,zhang2018fully} is the most popular one. Specifically, in UDA setting, by assuming that test or unlabeled data could be observed, the model is firstly trained on labeled source domain (\ie, training set) along with unlabeled target domain (\ie, test set), by reducing their domain gap. Then, the trained model is employed to segment the images from target domain. Nevertheless, these UDA based models require that target domain could be observed and even allowed to be trained. This prerequisite sometimes is difficult or infeasible to satisfy in the real application. For example, to protect personal privacy information, target domain (or test set) in some institutes cannot be directly accessed.

To alleviate the requirement of target domain in UDA, we consider a more feasible yet challenging setting, domain generalization (DG), to achieve generalizable medical image segmentation against domain shift. We notice that most existing DG models merely perform well in cross-central setting with small variations between domains, whereas the large domain shift (\eg, cross-modality) is seldom investigated that could largely deteriorate their performance~\cite{wang2020dofe,liu2020shape,liu2021feddg}.

We now illustrate two types of generalizable segmentation scenarios (\ie, cross-center and cross-modality) to clarify our motivation. Specifically, in Figure~\ref{fig1_b}, we show results of our method (denoted as ``Ours''), ``DeepAll'' baseline and a state-of-the-art cross-center DG method (``DoFE'')~\cite{wang2020dofe} on two different DG tasks. The first task is the \textbf{cross-center} prostate segmentation task~\cite{liu2020ms}. As illustrated in Figure~\ref{fig1_b}, all of these methods achieve relatively high Dice scores ($> 80\%$) on this task, and gaps of different methods are quite small. However, when we apply these three methods to BraTS dataset~\cite{menze2014multimodal}, a \textbf{cross-modality} brain tumor segmentation dataset, ``DeepAll'' and ``DoFE'' methods show a drastic degradation on Dice scores ($< 40\%$), while our method still achieves a competitive Dice score ($> 50\%$). The reason of this large performance degradation lies in large domain shift. For example, in Figure~\ref{fig1_a}, we illustrate T2- and T1-weighted images in BraTS dataset~\cite{menze2014multimodal}. The brain tumors that need to be segmented are delineated with red curves. It is obvious that T2 and T1 modalities show large distinct appearances. 
Accordingly, we notice that cross-modality DG task is more challenging than cross-center DG task due to the former should tackle larger domain shift. In this paper, we aim to deal with DG task with large domain shift (\eg, cross-modality task), and most previous DG methods are not designed for this.

Our setting owns its clinical meaning. For example, large distribution shift, caused by some unpredictable factors (\eg, interference from light source) during imaging, poses challenge to current generalization methods. Also, in some cases, data scarcity occurs in target domain makes UDA becomes hard to realize. In a nut shell, we intend to develop a robust method for realizing domain distribution shift insensitive modeling.

Based on above motivations, we propose a generalizable cross-modality medical image segmentation method trained on a single source domain (\eg, CT) and directly applied to unseen target domain (\eg, MRI) without any re-training. We notice that in medical images, modality discrepancy usually manifests in gray-scale distribution discrepancy. Being aware of this fact, we wish to simulate possible appearance changes in unseen target domains. In order to tackle this challenging cross-modality DG task, we introduce a module that can randomly augment source domain into different styles. To be specific, we utilize B\'ezier Curves~\cite{mortenson1999mathematics} as transformation functions to generate two groups of images: one group of images are similar to source domain images (\ie, \textit{source-similar domain}), and another group of images have a large distribution gap with source domain images (\ie, \textit{source-dissimilar domain}). Then, we introduce a segmentation model with a dual-normalization module to preserve style information of source-similar domain and source-dissimilar domain. Finally, a style-based path selection module is developed to help target domain images select the best normalization path to achieve optimal segmentation results. The main contributions of this paper are summarized as:

\begin{itemize}[leftmargin=6pt]
    \item We propose a deep dual-normalization model to tackle a more challenging DG task, \ie, generalizable cross-modality segmentation, that could directly segment the images from unseen target domains without re-training.
    \item We enhance the diversity of source domain via generating source-similar and source-dissimilar images based on B\'ezier Curves and develop a dual-normalization network for effective exploitation. Besides, we propose the style-based path selection scheme in the test stage.
    \item 
    Extensive experiments demonstrate our effectiveness. On BraTS dataset, our method achieves the Dice of 54.44\% and 57.98\% on T2 and T1CE source domains, respectively, which is quite close to UDA~\cite{chen2020unsupervised} (59.30\% on T2 source domain) as our upper bound . On both Cross-Modality Cardiac and Abdominal Multi-Organ datasets, our method outperforms the state-of-the-art DG methods.
\end{itemize}

\begin{figure*}[!htbp]
\centering
\includegraphics[width=1.0\textwidth]{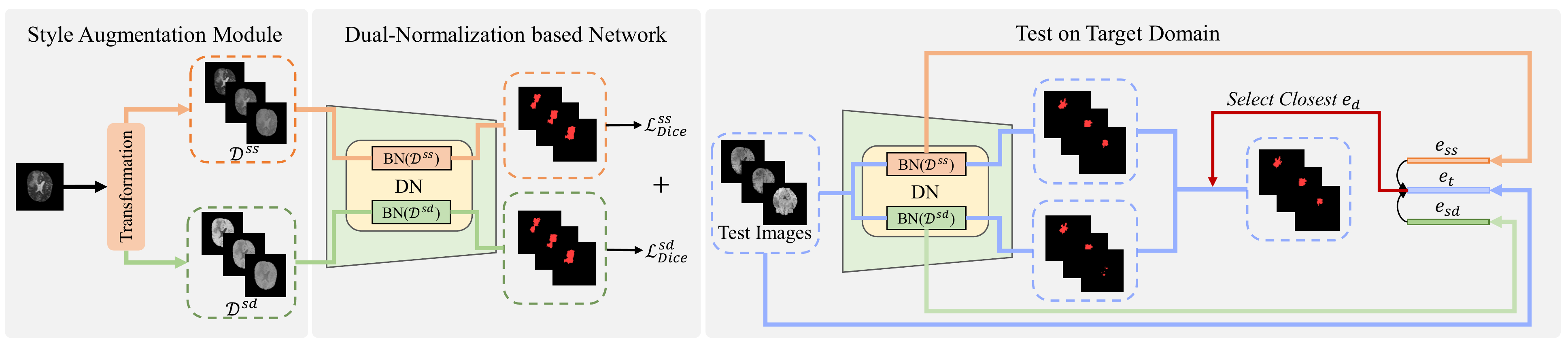}
\caption{The overall framework of our method. We first employ a style augmentation module to generate source domain into different styles and split them into source-similar domain ($\mathcal{D}^{ss}$) and source-dissimilar domain ($\mathcal{D}^{ds}$). Then, we train a dual-normalization (DN) segmentation network on ($\mathcal{D}^{ss}$) and ($\mathcal{D}^{sd}$). Finally, we test the trained network on target domains by style-based selection module.}
\label{fig3}
\vspace{-13pt}
\end{figure*}

\section{Related Work}

\paragraph{Unsupervised Domain Adaptation.}
The purpose of unsupervised domain adaptation (UDA) is to learn a model with labeled source domain and unlabeled target domain that retains promising performance on target domain~\cite{hoffman2018cycada,chen2019crdoco,zhang2018fully,chang2019all,chang2019domain,vu2019advent,liu2020open}. And UDA has attracted considerable attention recently. Some UDA methods utilize distribution alignment in pixel-level and adopt a generative network to narrow the domain gap between source and target domains~\cite{hoffman2018cycada,chen2019crdoco,zhang2018fully}. Chang \etal~\cite{chang2019all} use an adversarial training policy to align feature distribution between source and target domains to maintain semantic feature-level consistency across different domains. Also, Chang \etal~\cite{chang2019domain} separate batch normalization layers for each domain which allows the model to distinguish domain-specific and domain-invariant information. However, in some scenarios, \eg, the data could not be accessed during training due to privacy protection. Target domains are not available in the training process, causing UDA methods could not be directly uesd.
\vspace{-13pt}

\paragraph{Domain Generalization.}
Unlike UDA, domain generalization (DG), by training models purely on source domains, aims to directly generalize to target domains that could not be observed during the training process~\cite{zakharov2019deceptionnet,yue2019domain,qiao2020learning,gong2019dlow,du2020learning,chattopadhyay2020learning,hoffer2020augment}. Recently, a large number of works on DG tasks have been proposed. Among previous efforts, some methods are designed to learn domain-invariant representations by minimizing the domain discrepancy across multiple source domains~\cite{ghifary2015domain,hsu2017learning,li2018domain,li2018deep,motiian2017unified,zakharov2019deceptionnet,gong2021confidence}. Additionally, some methods use meta-learning, which employs episodic training policy by splitting source domain into meta-train and meta-test domains at each training iteration to simulate domain shift~\cite{li2018learning,balaji2018metareg,li2019episodic,dou2019domain,liu2020shape}. Besides, some methods tackle DG task by modifying normalization layers, \eg batch normalization (BN) and instance normalization (IN)~\cite{pan2018two,fan2021adversarially,seo2020learning,segu2020batch}. For example, Pan \etal~\cite{pan2018two} propose IBN-Net, which merges IN and BN layers in an unified framework, where IN could maintain invariant representation and BN is able to preserve discriminative features simultaneously. Also, Seo \etal~\cite{seo2020learning} introduce domain specific optimized normalization (DSON) to learn a joint embedding space across all source domains by optimizing domain-specific normalization layers. Segu \etal~\cite{segu2020batch} utilize domain-specific batch normalization layers to collect distribution statistics to model relations between features of source and target domains. There also exists some data augmentation approaches to increase the diversity of source domains for the sake of improving generalization ability on unseen target domains~\cite{volpi2018generalizing,zhang2020generalizing,zhou2020deep,zhou2020learning}. 

In medical image analysis, several previous works have studied generalizable medical image segmentation tasks. For example, Zhang \etal~\cite{zhang2020generalizing} propose a deep-stacked transformation approach that employs a series of transformations to simulate domain shift for a specific medical imaging modality. Wang \etal~\cite{wang2020dofe} build a domain knowledge pool to store domain-specific prior knowledge and then use domain attributes to aggregate features from different domains. Liu \etal~\cite{liu2021feddg} further improve the performance of cross-domain medical image segmentation by combining continuous frequency space interpolation with episodic training strategy. However, most existing DG methods for medical image segmentation work under small domain distribution shift. When large domain shift occurs, they might suffer performance degradation.

\section{Methodology}
\subsection{Definition and Overview}
We denote our single source domain as $\mathcal{D}^s = \{x_i^s, y_i^s\}_{i=1}^{N^s}$, where $s$ represents the domain ID, $x_i^s$ is the $i$-th image in the source domain $s$, $y_i^s$ is the segmentation mask of $x_i^s$, and $N^s$ is the total number of samples. Our purpose is to train a segmentation model $S_{\theta}: x \rightarrow y$ on source domain $\mathcal{D}^s$, where $x$ and $y$ represent the image set and label set in source domain $\mathcal{D}^s$, $S_{\theta}$ represents the segmentation model and $\theta$ is model parameters. We hope the model $S_{\theta}$ can generalize well to unseen target domain $\mathcal{D}^t$.

Specifically, we first propose a style augmentation module with several transformation functions to augment the source domain $\mathcal{D}^s$ into \textit{source-similar} domain $\mathcal{D}^{ss}$ and \textit{source-dissimilar} domain $\mathcal{D}^{sd}$. Then, based on the generated domain $\mathcal{D}^{ss}$ and $\mathcal{D}^{sd}$, a network equipped with a dual-normalization (DN) module is introduced in our method. We train the DN-based model on $\mathcal{D}^{ss}$ and $\mathcal{D}^{sd}$ domains. DN can preserve domain style information after model training. Finally, according to domain style information in DN and instance style information of the target domain, we can select the closest normalization statistics in DN to normalize features of target domain and get optimal segmentation results. The diagram of our method is shown in Figure~\ref{fig3}. We now discuss the technical details of our method.

\subsection{Style Augmentation Module}
For generalizable medical image segmentation task, using a single source domain to train the model is very tough. The style bias between different modalities will dramatically degrade the performance. From this perspective, we propose a simple yet effective style augmentation module to generate different stylized images from source domain.

Popular medical image modalities (\eg, X-ray, CT, and magnetic resonance images (MRI)) are usually gray-scale images. As is shown in Figure~\ref{fig1}, in T2-weighted MR brain images, whole tumor regions are much brighter than surroundings. In contrast, in T1-weighted MR brain images, foregrounds of whole tumors are darker than background regions. A simple idea to generate different styles is adjusting the gray value distribution of images. Inspired by the previous work Model Genesis~\cite{zhou2019models}, we adopt several monotonic non-linear transformation functions to map pixel values of original images to new values. Thus, the operation of changing gray distribution of images can be realized. Similar to~\cite{zhou2019models}, we use smooth and monotonic B\'ezier Curve~\cite{mortenson1999mathematics} as our transformation function.

\begin{figure}[t]
\centering
\includegraphics[width=1.0\columnwidth]{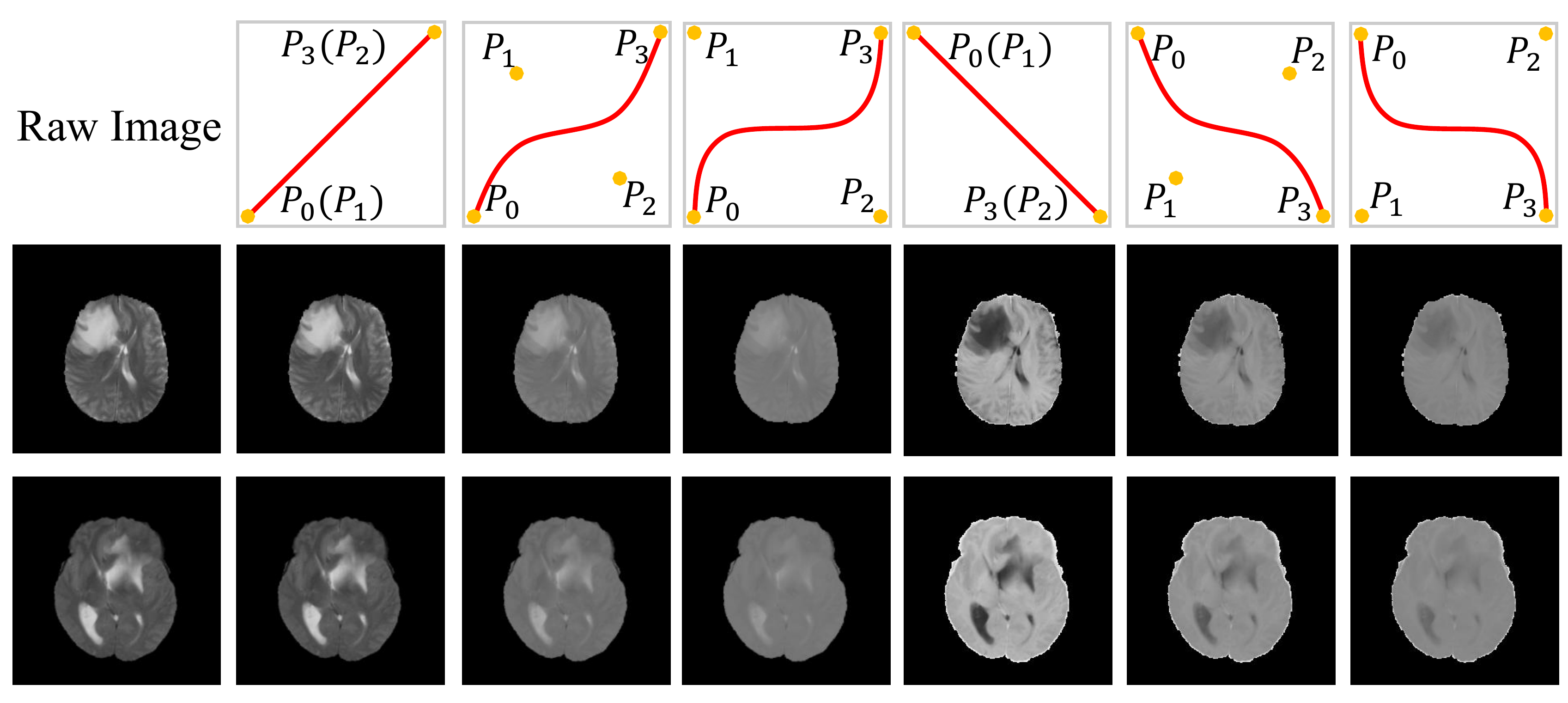}
\caption{T2 weight augmented MR Brain images and augmented images.}
\label{fig2}
\vspace{-15pt}
\end{figure}

B\'ezier Curve can be generated from two end points ($P_0$ and $P_3$) and two control points($P_1$ and $P_2$). The function is defined as follows:
\begin{displaymath}
    B(t) = \sum\limits_{i=0}^{n}\binom{n}{i}P_i(1-t)^{n-i}t^i, n = 3, t\in[0, 1],
\end{displaymath}
where $t$ is a fractional value along the length of the line. The domain and range of all B\'ezier Curves are $[-1, 1]$. In Figure~\ref{fig2}, we illustrate the original T2-weighted BraTS image and its augmented images. We set $P_0 = (-1, -1)$ and $P_3 = (1, 1)$ to get an increasing function and opposite to get a decreasing function. When $P_0 = P_1$ and $P_2 = P_3$, the B\'ezier Curve is a linear function (shown in Columns 2, 5). Then, we randomly generate another two pairs of control points. Specifically, we set $P_1 = (-v, v)$ and $P_2 = (v, -v)$ ($v \in (0, 1)$). We randomly generate two different $v$s for each image, so we get two increasing curves (shown in Columns 3 and 4) and two decreasing curves by inversion (shown in Columns 6, 7). Finally, we get 6 non-linear transformation functions (three increasing and three decreasing) for augmentation. In our three tasks, we normalize each sample to $[-1, 1]$. It should be noted that we only perform transformation operations on foreground regions.

Obviously, in gray-scale medical images, monotonically increasing transformation functions have less impact on image style. So we classify these transformed images obtained by increasing transformation functions as images similar to the source domain images, which we call \textit{source-similar} domain ($\mathcal{D}^{ss}$). On the contrary, these images generated by decreasing transformation functions will be treated as \textit{source-dissimilar} domain ($\mathcal{D}^{sd}$). We assume that those images with gray distribution close to source domain images can have good generalization performance on the model trained on $\mathcal{D}^{ss}$, while other images having large distribution gaps with source domain could generalize well on models trained on $\mathcal{D}^{sd}$. We use both domains to train the DN-based model introduced in the next section.

\subsection{Dual-normalization based Network}

It has been proven that batch normalization~\cite{ioffe2015batch} can make neural networks easy to capture data bias in their internal latent space~\cite{li2016revisiting}. However, data bias captured by neural networks with BN depends on domain distribution, which may degrade generalization ability when tested on novel domains. For our style-augmented images, simply adopting BN will make the model lose domain-specific distribution information of $\mathcal{D}^{ss}$ and $\mathcal{D}^{sd}$. As a result, our model may not be able to generalize well on target domains.

To capture different domain distribution information in $\mathcal{D}^{ss}$ and $\mathcal{D}^{sd}$, inspired by previous work~\cite{chang2019domain}, we adopt two different BN layers in a model to normalize activated features of $\mathcal{D}^{ss}$ and $\mathcal{D}^{sd}$, respectively. We call this Dual-Normalization (DN). The DN module can be written as
\begin{equation}
    \text{DN}(z; d)=\gamma_d\frac{z - \mu_d}{\sqrt{{\sigma_d}^2+\epsilon}} + \beta_d,
\end{equation}
where $z$ represents activated features of the model from domain $d$, $d$ represents the domain label, $\gamma_d$ and $\beta_d$ are affine parameters of domain $d$, $(\mu_d, {\sigma_d}^2)$ represent the mean and variance of the input feature $z$ from domain $d$ and $\epsilon > 0$ is a small constant value to avoid numerical instability.

During training, BN layers estimate means and variances of activated features by exponential moving average with update factor $\alpha$. For DN, they can be written as
\begin{equation}
    \overline{\mu}_d^{t+1} = (1 - \alpha)\overline{\mu}_d^{t} + \alpha \mu_d^{t},
\end{equation}
\begin{equation}
    (\overline{\sigma}_d^{t+1})^2 = (1 - \alpha)(\overline{\sigma}_d^{t})^2  + \alpha (\sigma_d^{t})^2,
\end{equation}
where $t$ represents current training iteration, $\overline{\mu}_d^{t}$ and $(\overline{\sigma}_d^{t})^2$ are the estimated means and variances of domain $d$ at $t$-th iteration. The estimated means and variances of DN for each domain can be considered as domain distribution information. When testing on target domain, these domain distribution information $(\mu_d, {\sigma_d}^2)$ can be used to compare with distribution information $(\mu_t, {\sigma_t}^2)$ of target domains. So that the model can select suitable domain distribution statistics to normalize activated features from target domains.

\subsection{Style-Based Path Selection}
The DN module allows the model to learn multiple source distribution of $\mathcal{D}^{ss}$ and $\mathcal{D}^{sd}$. So that the estimated statistics in DN can be considered as domain style information of $\mathcal{D}^{ss}$ and $\mathcal{D}^{sd}$. Hence, we get a light-weight ensemble model, where each domain shares same model parameters except for normalization statistics.

The model with DN module is trained on $\mathcal{D}^{ss}$ and $\mathcal{D}^{sd}$. After training, DN module will preserve statistics $\mu_d$ and $\sigma_d^2$ and affine parameters $\gamma_d$ and $\beta_d$ of training data from domain $d$. Therefore, we will get two series of $\mu_d$ and ${\sigma_d}^2$, which can be denoted as
\begin{displaymath}
    \begin{split}
        e_d =[e_d^1, e_d^2,...,e_d^L] =[(\mu_d^1, {\sigma_d^1}^2), (\mu_d^2, {\sigma_d^2}^2), ..., (\mu_d^L, {\sigma_d^L}^2)],
    \end{split}
\end{displaymath}
where $d$ represents the domain label of $\mathcal{D}^{ss}$ and $\mathcal{D}^{sd}$, the superscripts $l \in \{1, 2, ..., L\}$ denote the $l$-th BN layer in the model. This can be defined as a batch normalization embedding~\cite{segu2020batch} for a certain domain $d$. In our work, we denote $e_d$ as style embedding of domain $d$.

For a target sample $x_t$, we can capture instance statistics $(\mu_t, {\sigma_t}^2)$ by forward propagation. The style embedding $e_t$ of target domain sample can be described as
\begin{displaymath}
    \begin{split}
        e_t=[e_t^1, e_t^2,...,e_t^L]=[(\mu_t^1, {\sigma_t^1}^2), (\mu_t^2, {\sigma_t^2}^2), ..., (\mu_t^L, {\sigma_t^L}^2)].
    \end{split}
\end{displaymath}
Each $e_t^1$ represents instance style statistics of target domain sample at certain layer $l$ during forward propagation. Once the instance style embedding of target domain sample is available, it is possible to measure similarities of a target domain sample $x_t$ to $\mathcal{D}^{ss}$ and $\mathcal{D}^{sd}$ by calculating distances between $e_t$ and $e_d$.

To measure the distance between style embeddings of source domain and target domain, we adopt a symmetric distance function that satisfies the triangle inequality. In our method, we choose the Euclidean Distance. The distance of the $l$-th layer embeddings can be written as
\begin{equation}
    \begin{split}
        \mathcal{W}(e_t^l, e_d^l)&=\|\mu_t^l - \mu_d^l\|_2^2 + \|{\sigma_t^l}^2 - {\sigma_d^l}^2\|_2^2.
    \end{split}
\end{equation}
We measure the distance between target sample $x_t$ and source domain $d$ by summing over the distance between the style embeddings $e_t^l$ and $e_d^l$ of all layers:
\begin{equation}
    \begin{split}
        \text{Dist}(e_t, e_d)&=  \sum\limits_{l \in \{1, 2, ...,L\}}\mathcal{W}(e_t^l, e_d^l).
    \end{split}
\end{equation}

Once the distance to each source domain is computed, we can choose the nearest source domain style embedding and affine parameters $\gamma_d$ and $\beta_d$ to normalize the input feature $z_t$ of target domain:
\begin{equation}
    \begin{split}
        c = \mathop{\arg\min}\limits_{d} \text{Dist}(e_t, e_d).
    \end{split}
\end{equation}
The normalization on target domain feature $z_t$ is written as 
\begin{equation}
\label{eq}
    \text{Norm}(z_t; c)=\gamma_c\frac{z_t - \mu_c}{\sqrt{{\sigma_c}^2+\epsilon}} + \beta_c.
\end{equation}

Since our model shares all parameters except batch normalization layers on $\mathcal{D}^{ss}$ and $\mathcal{D}^{sd}$, the model we trained can predict results on target domains by normalized features in Equation (\ref{eq}). We denote $S_{\theta}(\cdot)$ as our segmentation model, where $\theta$ represents parameters in the model except for batch normalization layers. So we can form predicting result on target domain $t$ as $S_{\theta}(z_t^c)$, where $z_t^c$ represents normalized target domain features from Equation (\ref{eq}).

\subsection{Training Details}
As aforementioned, DN module contains two separate batch normalization layers in our model---one is for $\mathcal{D}^{ss}$, and another is for $\mathcal{D}^{sd}$. At the beginning, we augment the source domain into $\mathcal{D}^{ss}$ and $\mathcal{D}^{sd}$ by style augmentation module. Then, we feed them in the DN-based model to get soft predictions. Afterwards, we optimize the model $S_{\theta}$ by segmentation loss. To overcome the class imbalance issue between relative small-sized foreground and large-sized background, we employ a sum of soft Dice loss of $\mathcal{D}^{ss}$ and $\mathcal{D}^{sd}$ to train the segmentation network:
\begin{equation}
        \mathcal{L}_{seg} = \mathcal{L}_{Dice}(S_{\theta}(x_{ss}), y_{ss}) +
        \mathcal{L}_{Dice}(S_{\theta}(x_{sd}), y_{sd}),
\end{equation}
where $(x_{ss}, y_{ss})$ and $(x_{sd}, y_{sd})$ represent pairs of image and related one-hot ground truth from $\mathcal{D}^{ss}$ and $\mathcal{D}^{sd}$, $S_{\theta}(\cdot)$ yields soft prediction. We show the overall framework of our method in Figure~\ref{fig3}.

\section{Experiments}

\begin{table*}[t]
\centering
\caption{Comparison of different methods on the BraTS dataset. $\uparrow$: the higher the better, $\downarrow$: the lower the better.}
\scriptsize{
\setlength{\tabcolsep}{1mm}{
\begin{tabular}{c|cccc|cccc|cccc|cccc}
    \toprule[0.8pt]
    & \multicolumn{8}{c|}{Source Domain: T2} & \multicolumn{8}{c}{Source Domain: T1CE} \\
    \midrule[0.5pt]
    \multirow{2}{*}{Method} & \multicolumn{4}{c|}{Dice (\%) $\uparrow$} & \multicolumn{4}{c|}{Hausdorff Distance (mm) $\downarrow$} &\multicolumn{4}{c|}{Dice (\%) $\uparrow$} & \multicolumn{4}{c}{Hausdorff Distance (mm) $\downarrow$}\\
    \cmidrule[0.5pt]{2-17}
    & Flair & T1 & T1CE & Average & Flair & T1 & T1CE & Average & Flair & T1 & T2 & Average & Flair & T1 & T2 & Average \\
    \midrule[0.5pt]
    No Adaptation & 70.01 & 5.58 & 9.33 & 28.31 & 20.52 & 56.51 & 50.03 & 42.35 & 37.53 & 59.13 & 11.13 & 35.93 & 26.32 & 18.97 & 50.23 & 31.84 \\
    \midrule[0.5pt]
    Source-Similar & 71.49 & 5.83 & 8.87 & 28.73 & 20.46 & 57.01 & 56.29 & 44.59 & 45.58 & 62.56 & 18.66 & 42.27 & 21.06 & 19.60 & 50.21 & 30.29 \\
    Source-Dissimilar & 0.48 & 47.48 & 36.52 & 28.16 & 60.28 & 22.35 & 22.84 & 35.16 & 15.68 & 5.54 & 65.87 & 29.03 & 55.69 & 58.41 & 17.02 & 43.71\\
    \midrule[0.5pt]
    DeepAll & \textbf{77.44} & 13.65 & 14.42 & 35.17 & 13.54 & 35.06 & 30.28 & 26.29 & 38.48 & 48.67 & 19.26 & 35.47 & 26.79 & 19.65 & 48.21 & 31.55\\
    IBN-Net~\cite{pan2018two} & 76.56 & 5.41 & 6.77 & 29.58 & \textbf{13.02} & 55.75 & 51.28 & 39.96 & \textbf{50.23} & 46.66 & 15.52 & 37.47 & 21.56 & 22.98 & 50.67 & 31.74\\
    DSON~\cite{seo2020learning} & 75.69 & 5.75 & 9.90 & 30.45 & 25.23 & 34.68 & 35.28 & 31.73 & 55.60 & 59.44 & 13.40 & 42.81 & 30.50 & 29.91 & 36.06 & 32.16\\
    MLDG~\cite{li2018learning} & 71.23 & 5.47 & 8.83 & 28.51 & 15.64 & 56.02 & 51.01 & 40.89 & 29.53 & 51.38 & 3.56 & 28.16 & 32.84 & 23.06 & 55.39 & 37.10\\
    DoFE~\cite{wang2020dofe} & 74.91 & 5.72 & 9.31 & 29.98 & 14.18 & 55.64 & 50.43 & 40.08 & 32.25 & 56.82 & 4.12 & 31.06 & 31.66 & 21.08 & 56.69 & 36.48\\
    Fed-DG~\cite{liu2021feddg} & 75.77 & 5.82 & 9.51 & 30.37 & 14.45 & 54.03 & 51.06 & 39.85 & 33.03 & 58.30 & 4.09 & 31.72 & 32.07 & 22.35 & 56.08 & 36.83\\
    \textbf{Ours} & 75.87 & \textbf{49.36} & \textbf{38.09} & \textbf{54.44} & 13.44 & \textbf{20.15} & \textbf{23.56} & \textbf{19.05} & 47.31 & \textbf{63.64} & \textbf{63.00} & \textbf{57.98} & \textbf{21.03} & \textbf{18.06} & \textbf{17.56} & \textbf{18.88}\\
    \bottomrule[0.8pt]
\end{tabular}
}
}
\label{table1}
\end{table*}

\begin{table*}[t]
\centering
\caption{Comparison of different methods on the Cardiac dataset. $\uparrow$: the higher the better, $\downarrow$: the lower the better.}
\scriptsize{
\setlength{\tabcolsep}{0.8mm}{
\begin{tabular}{c|ccccc|ccccc|ccccc|ccccc}
    \toprule[0.8pt]
    & \multicolumn{10}{c|}{Cardiac MRI $\rightarrow$ Cardiac CT} & \multicolumn{10}{c}{Cardiac CT $\rightarrow$ Cardiac MRI}\\
    \midrule[0.5pt]
    \multirow{2}{*}{Method} & \multicolumn{5}{c|}{Dice (\%) $\uparrow$} & \multicolumn{5}{c|}{Hausdorff Distance (mm) $\downarrow$} & \multicolumn{5}{c|}{Dice (\%) $\uparrow$} & \multicolumn{5}{c}{Hausdorff Distance (mm) $\downarrow$} \\
    \cmidrule[0.5pt]{2-21}
    & AA & LAC & LVC & MYO & Average & AA & LAC & LVC & MYO & Average & AA & LAC & LVC & MYO & Average & AA & LAC & LVC & MYO & Average \\
    \midrule[0.5pt]
    No Adaptation & 27.10 & 28.92 & 2.24 & 1.88 & 15.04 & 87.06 & 86.09 & 84.71 & 84.64 & 85.62 & 4.61 & 3.91 & 3.94 & 4.52 & 4.24 & 101.08 & 100.53 & 100.22 & 101.30 & 100.78 \\
    \midrule[0.5pt]
    DeepAll & 39.07 & 38.65 & 41.56 & 41.17 & 40.11 & 26.51 & 30.05 & 28.41 & 25.03 & 27.50 & 18.12 & 18.04 & 19.44 & 17.71 & 18.33 & 108.92 & 105.05 & 108.55 & 111.69 & 108.55 \\
    IBN-Net~\cite{pan2018two} & 28.48 & 25.28 & 32.31 & 28.43 & 28.63 & 80.64 & 82.17 & 76.38 & 75.65 & 78.71 & 25.79 & 24.98 & 24.71 & 26.10 & 25.39 & 61.42 & \textbf{62.12} & 69.85 & 79.24 & 68.16 \\
    DSON~\cite{seo2020learning} & 43.09 & 40.24 & 42.66 & 42.00 & 42.00 & 26.56 & 28.51 & 22.73 & 25.21 & 25.75 & 26.13 & 27.21 & 24.75 & 21.52 & 24.90 & 78.86 & 77.20 & 78.20 & 89.67 & 80.98 \\
    MLDG~\cite{li2018learning} & 50.34 & 48.30 & 46.69 & 45.49 & 47.71 & 17.06 & 20.64 & 18.43 & 17.68 & 18.45 & 23.65 & 28.31 & 25.62 & 20.79 & 24.59 & 75.64 & 70.31 & 69.58 & 71.46 & 71.75 \\
    DoFE~\cite{wang2020dofe} & 51.38 & 48.51 & 46.47 & 44.99 & 47.84 & \textbf{15.20} & 17.02 & 16.38 & 18.22 & 16.70 & 25.96 & 25.87 & 25.49 & 22.96 & 25.07 & 77.25 & 68.01 & 75.41 & 79.06 & 74.93 \\
    Fed-DG~\cite{liu2021feddg} & \textbf{52.41} & 49.14 & 46.77 & 43.93 & 48.06 & 16.83 & 21.14 & 19.95 & 20.36 & 19.57 & 25.47 & 28.08 & 24.25 & 22.21 & 25.00 & 80.29 & 82.54 & 79.68 & 72.51 & 78.76 \\
    \textbf{Ours} & 51.42 & \textbf{50.20} & \textbf{52.86} & \textbf{52.31} & \textbf{51.70} & 16.14 & \textbf{16.76} & \textbf{15.11} & \textbf{16.20} & \textbf{16.05} & \textbf{33.38} & \textbf{31.65} & \textbf{33.29} & \textbf{30.45} & \textbf{32.19} & \textbf{59.04} & 69.01 & \textbf{68.24} & \textbf{66.03} & \textbf{65.58} \\
    \bottomrule[0.8pt]
\end{tabular}
}
}
\label{table2}
\vspace{-12pt}
\end{table*}

\subsection{Experimental Setting}
\textbf{Datasets and Preprocessing.}
We introduce three datasets (\ie, BraTS dataset~\cite{menze2014multimodal}, Cross-Modality Cardiac dataset~\cite{zhuang2016multi} and Abdominal Multi-Organ datasets~\cite{kavur2021chaos,landman2015miccai}) for evaluation. The Cross-Modality Brain Tumor Segmentation Challenge 2018 dataset (BraTS)~\cite{menze2014multimodal} is composed of four modalities of MR images, \ie, T2, Flair, T1, and T1CE. 

Cross-Modality Cardiac dataset~\cite{zhuang2016multi} consists of 20 unpaired MRI and CT volumes from different clinical sites, which contains four cardiac structures, \ie, left ventricle myocardium (LVM), left atrium blood cavity (LAB), left ventricle blood cavity (LVB), and ascending aorta (AA). 

The Abdominal Multi-Organ dataset contains two different modalities. One is the T2-SPIR MRI training data from ISBI 2019 CHAOS Challenge~\cite{kavur2021chaos} with 20 volumes. The other one is the publicly available CT data from~\cite{landman2015miccai} with 30 volumes. This dataset contains four abdominal organs, \ie, liver, right kidney (R. Kid), left kidney (L. Kid), and spleen. We use the manual delineation of these datasets provided by professional radiologists as ground truth for evaluation.

For image preprocessing, we normalize the image to [-1, 1] in intensity values. For the Abdominal Multi-Organ dataset, we crop the volume of each case that contains segmentation targets. Following the previous work in UDA~\cite{zou2020unsupervised}, we randomly select 80\% of patient data as training set and 20\% as test set, and each slice is resized to $256 \times 256$. During training, we perform data augmentations \ie, random crop, random rotation, random scale, \etc.

\textbf{Implementation Details.}
We employ U-Net~\cite{ronneberger2015u} as our segmentation backbone with replacing all BN layers to our DN module. We implement our model with the PyTorch framework on 4 Nvidia RTX 2080Ti GPUs with 11 GB memory. For all datasets, we train the model for 50 epochs with a batch size of 64. We choose the Adam optimizer with an initial learning rate $lr_0$ of $4 \times 10^{-3}$ as our optimizer to train the model. Additionally, for stable training, the learning rate $lr$ is decayed according to the polynomial rule.

\textbf{Evaluation Metrics.}
We adopt two popularly used evaluation metrics, \ie, Dice coefficient (Dice) and Hausdorff distance (HD). The Dice coefficient measures the overlapping ratio between prediction and ground truth. The higher Dice value, the better segmentation performance. Hausdorff distance is defined between two sets in the metric space. The lower HD value, the better performance. 

\subsection{Comparison with State-of-the-art Methods}

\begin{table*}[h]
\centering
\caption{Comparison of different methods on the Abdominal Multi-Organ dataset. $\uparrow$: the higher the better, $\downarrow$: the lower the better.}
\scriptsize{
\setlength{\tabcolsep}{0.65mm}{
\begin{tabular}{c|ccccc|ccccc|ccccc|ccccc}
    \toprule[0.8pt]
    & \multicolumn{10}{c|}{Abdominal MRI $\rightarrow$ Abdominal CT} & \multicolumn{10}{c}{Abdominal CT $\rightarrow$ Abdominal MRI}\\
    \midrule[0.5pt]
    \multirow{2}{*}{Method} & \multicolumn{5}{c|}{Dice (\%) $\uparrow$} & \multicolumn{5}{c|}{Hausdorff Distance (mm) $\downarrow$} & \multicolumn{5}{c|}{Dice (\%) $\uparrow$} & \multicolumn{5}{c}{Hausdorff Distance (mm) $\downarrow$} \\
    \cmidrule[0.5pt]{2-21}
    & Spleen & R. Kid & L. Kid & Liver & Average & Spleen & R. Kid & L. Kid & Liver & Average & Spleen & R. Kid & L. Kid & Liver & Average & Spleen & R. Kid & L. Kid & Liver & Average \\
    \midrule[0.5pt]
    No Adaptation & 7.98 & 6.53 & 7.60 & 7.21 & 7.33 & 53.20 & 49.91 & 52.81 & 47.53 & 50.86 & 6.16 & 6.03 & 3.37 & 2.80 & 4.59 & 35.97 & 44.23 & 26.00 & 33.85 & 35.01  \\
    \midrule[0.5pt]
    DeepAll & 17.37 & 17.86 & 16.56 & 18.05 & 17.46 & 36.37 & 38.74 & 37.42 & 37.02 & 37.39 & 22.81 & 26.49 & 21.02 & 23.43 & 23.44 & 28.65 & 17.95 & 10.33 & 16.61 & 18.39 \\
    IBN-Net~\cite{pan2018two} & 11.11 & 12.50 & 15.67 & 14.96 & 13.56 & 40.21 & 42.31 & 39.06 & 38.52 & 40.03 & 19.31 & 25.08 & 22.65 & 27.14 & 23.55 & 30.05 & 20.16 & 15.67 & 19.08 & 21.24 \\
    DSON~\cite{seo2020learning} & 7.12 & 7.98 & 10.06 & 9.26 & 8.61 & 45.17 & 40.36 & 42.13 & 49.33 & 44.25 & 8.05 & 18.60 & 15.72 & 8.50 & 12.72 & 46.55 & 9.65 & 16.58 & 19.72 & 23.12 \\
    MLDG~\cite{li2018learning} & 31.89 & 34.21 & 34.88 & 37.85 & 34.71 & 29.40 & 26.13 & 27.88 & 25.03 & 27.11 & 41.05 & 37.44 & 35.82 & 39.46 & 38.44 & 25.61 & 12.76 & 12.06 & 12.12 & 15.64 \\
    DoFE~\cite{wang2020dofe} & 33.18 & 37.33 & 36.20 & 44.67 & 37.85 & 18.58 & 12.18 & 17.24 & 17.81 & 16.45 & 41.36 & 36.97 & 36.47 & 39.55 & 38.59 & 25.52 & 9.98 & 11.69 & 16.92 & 16.03 \\
    Fed-DG~\cite{liu2021feddg} & 32.54 & 36.15 & \textbf{41.12} & \textbf{47.06} & 39.22 & 12.75 & 9.78 & 12.90 & \textbf{9.61} & 11.26 & 21.48 & 18.54 & 57.40 & 58.22 & 38.91 & 21.94 & 27.73 & 10.51 & 6.50 & 16.67 \\
    \textbf{Ours} & \textbf{37.53} & \textbf{37.87} & 40.94 & 42.14 & \textbf{39.62} & \textbf{11.45} & \textbf{8.81} & \textbf{10.44} & 9.68 & \textbf{10.10} & \textbf{68.36} & \textbf{71.54} & \textbf{73.70} & \textbf{67.27} & \textbf{70.22} & \textbf{6.40} & \textbf{2.00} & \textbf{1.37} & \textbf{1.99} & \textbf{2.94} \\
    \bottomrule[0.8pt]
\end{tabular}
}
}
\label{table3}
\vspace{-15pt}
\end{table*}

In Table~\ref{table1}, we report the results of source domain T2 and T1CE in BraTS dataset. ``No Adaptation'' indicates the results of target domains by directly applying the model trained on single source domain. Besides, we report the results of models trained on $\mathcal{D}^{ss}$ and $\mathcal{D}^{sd}$, which are described as ``Source-Similar'' and ``Source-Dissimilar''. Several recently proposed SOTA methods are also included. First, ``DeepAll'' (\ie, directly training on aggregated source domains and testing on target domains) is regarded as the baseline in our evaluation. Moreover, we choose five DG methods for comparison, including IBN-Net~\cite{pan2018two} and DSON~\cite{seo2020learning}: two normalization based methods, MLDG~\cite{li2018learning} and Fed-DG~\cite{liu2021feddg}: two meta-learning based methods, and DoFE~\cite{wang2020dofe}: a domain-invariant feature learning approach. These DG methods mentioned above all focus on learning or keeping domain invariant information, while our method focuses on selecting the most similar domain information to help generalization on target domains.

As we expected, our method outperforms all other methods on average results of Dice score and HD with huge margins. Specifically, on source domain T2, our method achieves the highest average Dice score of 54.44\% and the lowest average HD of 19.05 mm. Compared to the best results of other approaches (\vs ``DeepAll'' ), we increased the Dice score by a large margin of 19.27\%. Similarly, on source domain T1CE, our method improves the Dice score by 20.51\% compared to the best SOTA result (\vs ``IBN-Net''). We also observe an interesting fact that on source domain T2, the average results of baseline ``DeepAll'' surpass all other DG methods. This is because all of the compared DG methods mainly focus on addressing DG tasks with small distribution shift (\eg, cross-center tasks) when applying to DG tasks with large distribution shift (\eg, cross-modality tasks), their performance will deteriorate a lot. We believe the superior performance can be attributed to the fact that, with the DN-based model and style-based selection, our method is capable of generalizing well on target domains in cross-modality DG tasks. 

Additionally, we notice that a single segmentation model trained on $\mathcal{D}^{ss}$ and $\mathcal{D}^{ss}$ (\ie, ``Source-Similar'' and ``Source-Dissimilar'' in Table~\ref{table1}) separately could not guarantee generalization quality on all target domains. Taking source domain T2, for example, in ``Source-Similar'', the model gets a Dice score of 71.49\% on target domain Flair. However, when testing on target domain T1 and T1CE, the Dice scores are only 5.58\% and 8.87\%. Similarly, in ``Source-Dissimilar'', the Dice score on Flair is only 0.48\%. This indicates that separately trained on $\mathcal{D}^{ss}$ or $\mathcal{D}^{sd}$, the model can not generalize well on all target domains.

The segmentation performance of all methods on the Cardiac dataset and Abdominal Multi-Organ dataset is given in Table~\ref{table2} and Table~\ref{table3}. In both experiments, the average performance of ``No Adaptation'' is surprisingly worse than all other methods. This reveals that without adaptation or generalization techniques, the model is unable to generalize on target domains. Furthermore, we still notice that the baseline ``DeepAll'' can outperform some other well-designed DG methods in both tasks, which further illustrates that most DG methods are not suitable for DG tasks with large domain shift. Given that our method is specialized in dealing with the cross-modality DG task, we achieve stable performance gain on both segmentation tasks.

\begin{figure}[t]
\centering
\includegraphics[width=1.0\columnwidth]{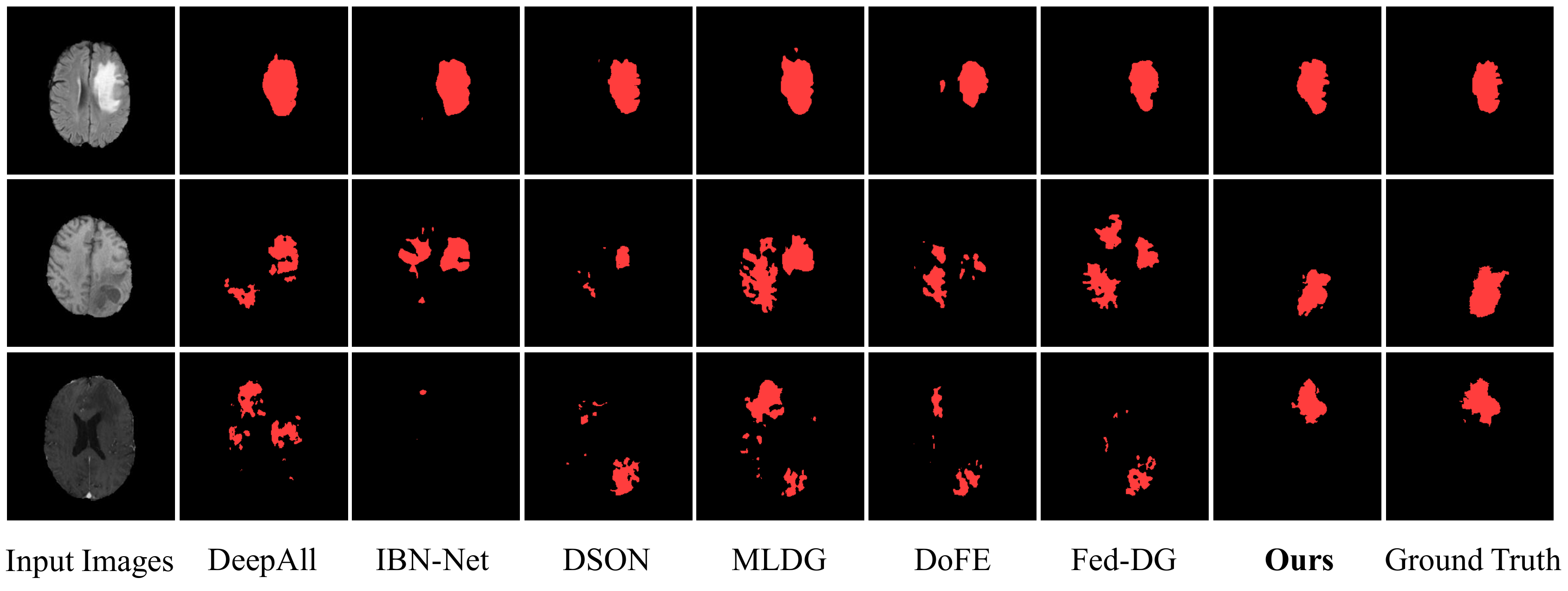}
\caption{Visualization results (\ie, Flair, T1 and T1CE) of BraTS dataset obtained by our method and other methods trained on source domain T2, together with the ground truth. }
\label{fig4}
\vspace{-8pt}
\end{figure}

\begin{figure}[t]
\centering
\includegraphics[width=1.0\columnwidth]{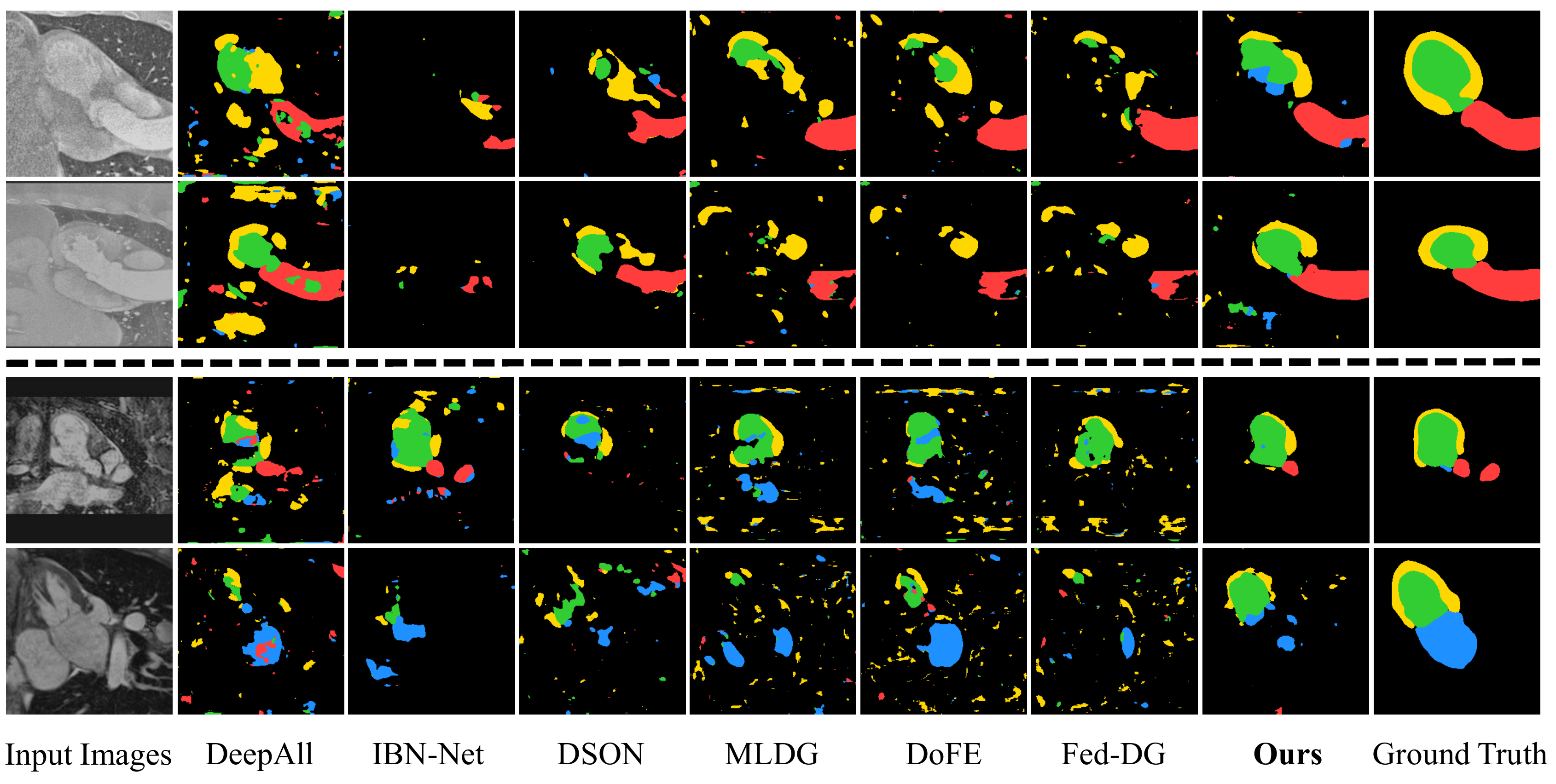}
\caption{Visualization results on Cardiac segmentation task of different methods. First two rows: ``MR to CT'' task; Last two rows: ``CT to MR'' task.}
\label{fig5}
\vspace{-14pt}
\end{figure}

\begin{figure}[t]
\centering
\includegraphics[width=1.0\columnwidth]{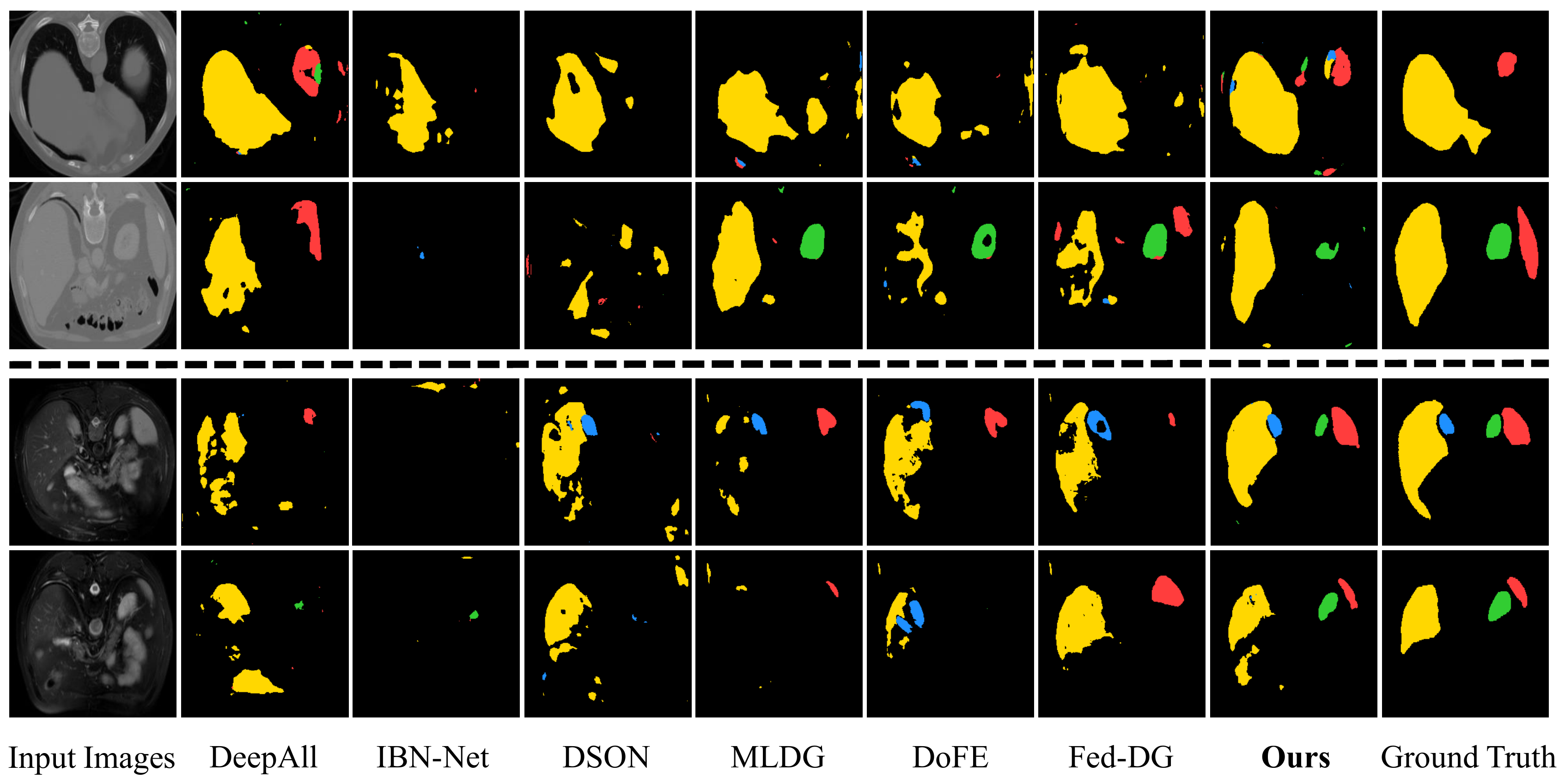} 
\caption{Visualization results on Multi-Organ segmentation task of different methods. First two rows: ``MR to CT'' task; Last two rows: ``CT to MR'' task.}
\label{fig6}
\vspace{-10pt}
\end{figure}

We visualize the segmentation results of our method and other methods on three tasks in Figure~\ref{fig4},~\ref{fig5} and~\ref{fig6}. They show that our model can produce more accurate segmentation results of target domains, especially having good spatial continuity of segmentation targets.

\subsection{Discussion}
\begin{table}[t]
\centering
\caption{Results of our model without style-based selection on BraTS dataset.}
\scriptsize{
\begin{tabular}{c|cccc|c}
 \toprule[0.8pt]
 \multicolumn{6}{c}{Source-Similar BN Path}\\
 \midrule[0.5pt]
 Domain & T2 & Flair & T1CE & T1 & Average \\
 \midrule[0.5pt]
 T2 & 82.52 & \underline{75.87} & \underline{9.58} & \underline{6.48} & 30.64 \\
 T1CE & \underline{14.48} & \underline{47.31} & 73.50 & \underline{63.64} & 62.72 \\
 \midrule[0.8pt]
 \multicolumn{6}{c}{Source-Dissimilar BN Path}\\
 \midrule[0.5pt]
 Domain & T2 & Flair & T1CE & T1 & Average \\
 \midrule[0.5pt]
 T2 & 1.42 & \underline{0.94} & \underline{40.17} & \underline{49.36} & 30.16 \\
 T1CE & \underline{63.00} & \underline{23.57} & 3.83 & \underline{7.34} & 31.30 \\
 \bottomrule[0.8pt]
\end{tabular}
}
\label{table4}
\vspace{-12pt}
\end{table}

\begin{figure}[t]
\centering
\includegraphics[width=0.90\columnwidth]{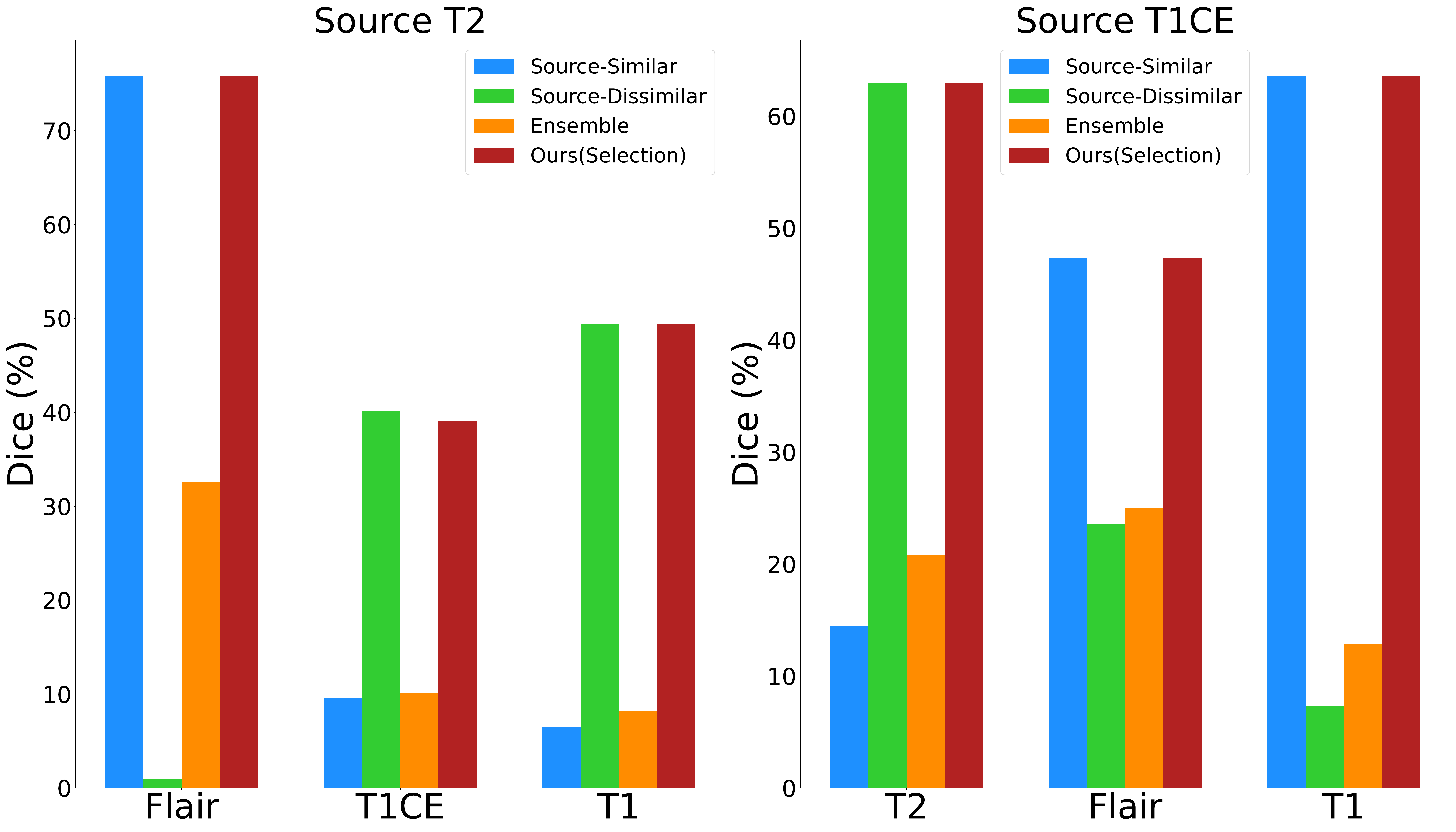} 
\caption{Dice score comparison of style-based selection method, non-selection method and ensemble policy on BraTS dataset.}
\label{fig7}
\end{figure}

\paragraph{Efficacy of Style-based Path Selection.} Since our model involves the DN module, we wish to validate the effectiveness of our style-based path selection module. Specifically, on BraTS dataset, we evaluate predictions obtained by source-similar BN path and source-dissimilar BN path on all domains (including source domain). As reported in Table~\ref{table4}, each row represents modality of source domain, and each column represents modality of tested domain. We calculate average Dice scores of all target domains (underlined results in the table). It shows that neither source-similar BN path nor source-dissimilar BN path can generalize well on all target domains. Additionally, we display results of our style-based selection method with source-similar BN path and source-dissimilar BN path in Figure~\ref{fig7}. It reveals that our method is more robust. Also, we ensemble the results produced by source-similar BN path and source-dissimilar BN path. The blue, green, orange and red bars represent results of source-similar BN, source-dissimilar BN, ensemble predictions, and style-based selection module, respectively. It indicates that ensemble predictions cannot receive promising results, and our style-based selection could help select relative optimal results.

\begin{figure}[t]
    \centering
    \begin{subfigure}{0.45\linewidth}
        \centering
        \includegraphics[width=1.0\columnwidth]{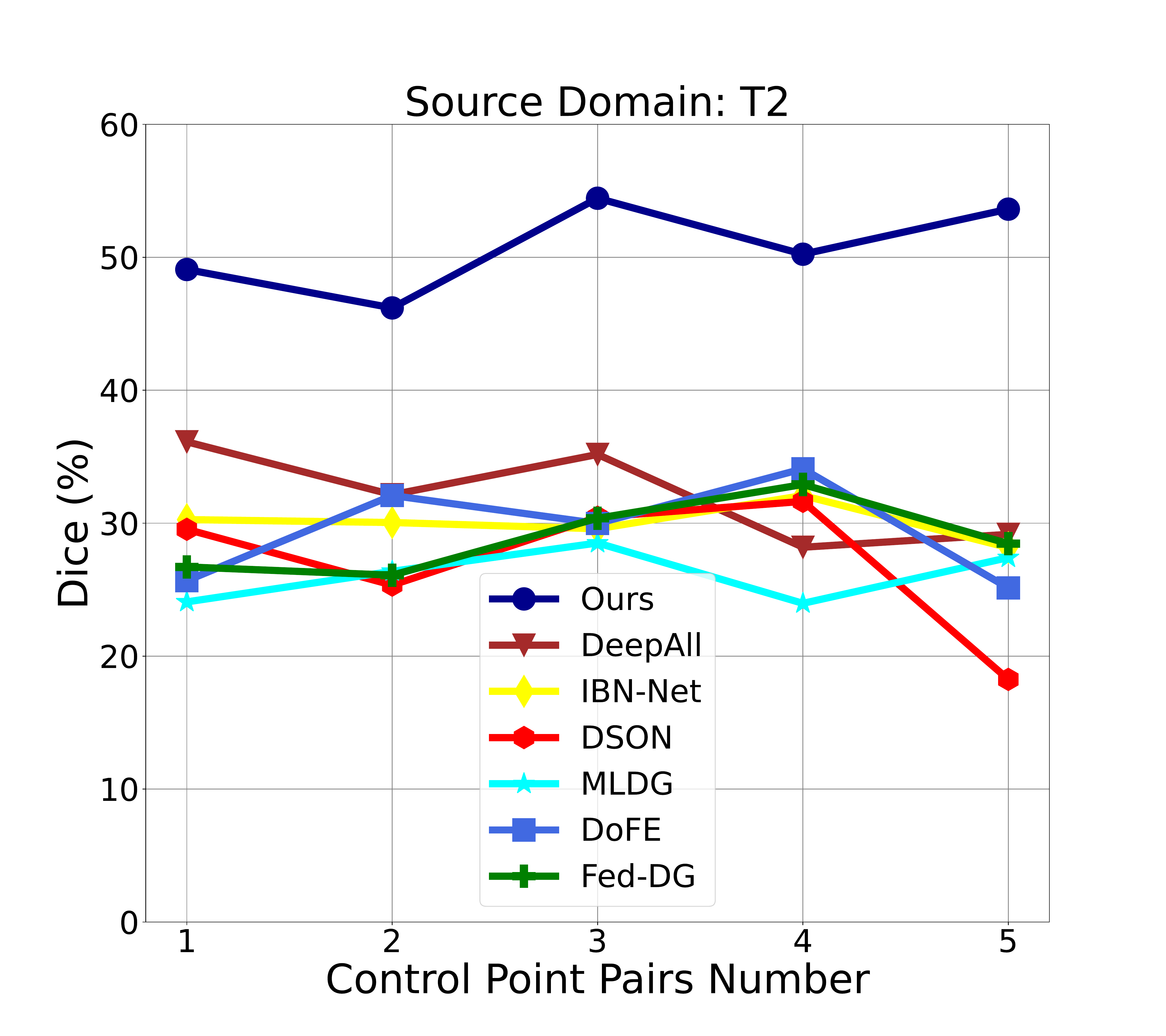}
        \caption{Source Domain T2}
        \label{fig8_a}
    \end{subfigure}
    \begin{subfigure}{0.45\linewidth}
        \centering
        \includegraphics[width=1.0\columnwidth]{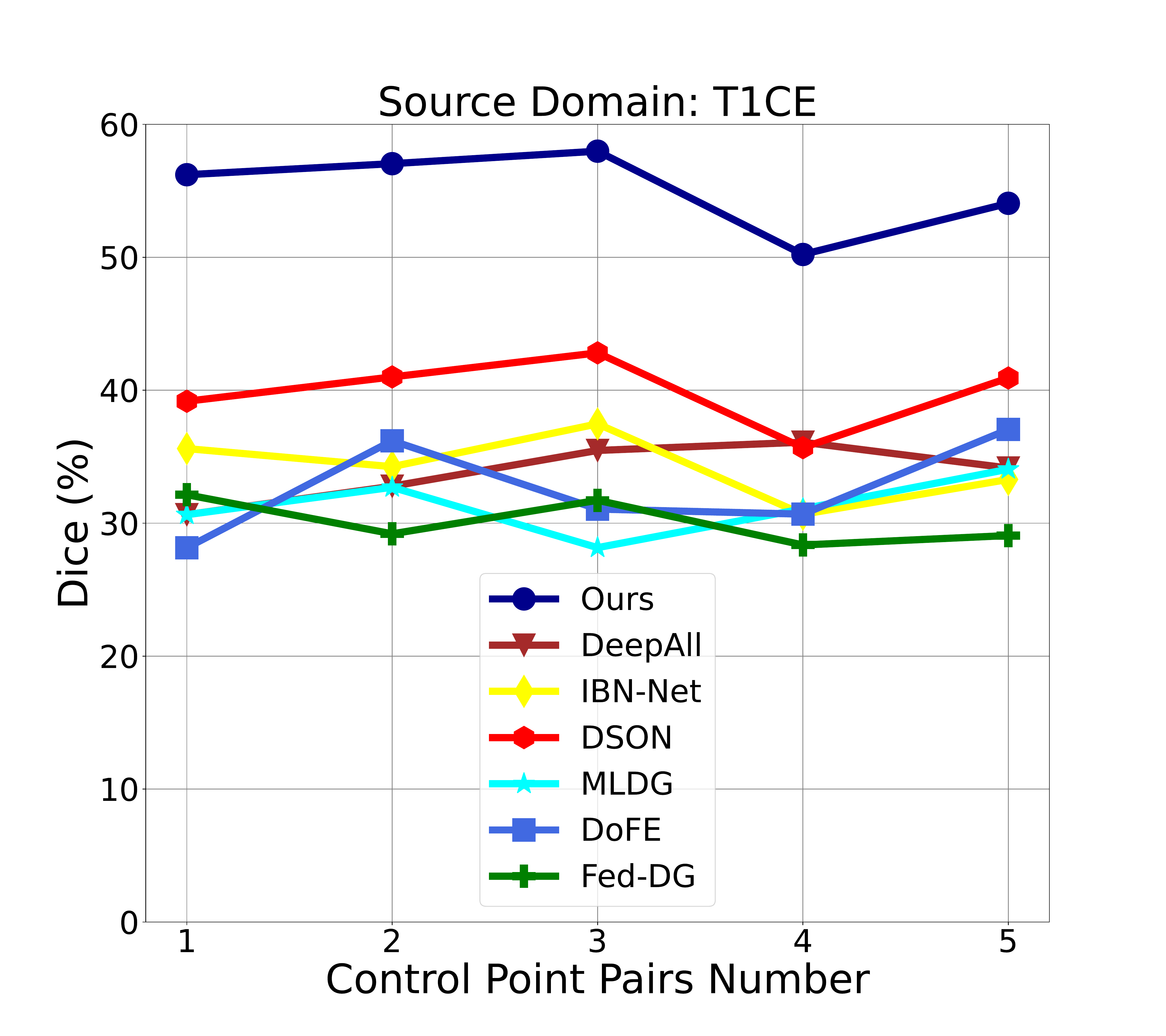}
        \caption{Source Domain T1CE}
        \label{fig8_b}
    \end{subfigure}
    \caption{The segmentation performance on BraTS dataset of our method and other SOTA methods based on different numbers of control point pairs.}
    \label{fig8}
    \vspace{-15pt}
\end{figure}

\vspace{-12pt}
\paragraph{Efficacy of Style Augmentation.}
In Section 3.1, we randomly generate three images as source-similar and three images as source-dissimilar in each case by using B\'ezier Curves. The number of pairs of control points B\'ezier Curve is proportional to the number of augmented images, which will contribute to more training time. We conduct an ablation study to analyze how the number of transformation functions pairs will influence results. So, we explore different numbers of control point pairs from 1 to 5 in Figure~\ref{fig8}. The vertical axis represents the Dice score and the horizontal axis represents the number of functions. We observe that, regardless of the number of transformation functions, segmentation results of our method exceed other methods by a large margin. This proves that the number of transformation functions will not affect the results of our method much when controlled in a reasonable range. 

\vspace{-12pt}
\paragraph{Analysis on Cross-center Task.}
To further verify the performance of our method on cross-center DG tasks, we evaluate our method on cross-center prostate segmentation~\cite{liu2020ms}. This is a well-organized cross-center dataset for prostate MRI segmentation. In this task, the Dice score of our method is 84.19\%, and the baseline (``DeepAll'') and SOTA (``DoFE''~\cite{wang2020dofe}) methods obtain Dice scores of 85.46\% and 87.48\% (reported in~\cite{wang2020dofe}), respectively. Although our method does not outperform others in cross-center tasks, their gaps are relatively small. We need to mention that our approach aims to solve DG tasks with large domain shift (\eg, cross-modality task), and results in Section 4.2 also prove that our method shows huge advantages on three datasets of cross-modality DG tasks.

\section{Conclusion}
In this paper, we first attempt to study the generalizable cross-modality medical image segmentation task. We employ B\'ezier Curves to augment single source domain $\mathcal{D}^{s}$ into different styles and split them into source-similar domain $\mathcal{D}^{ss}$ and source-dissimilar domain $\mathcal{D}^{sd}$. Moreover, we design a dual-normalization module to estimate domain distribution information. During the test stage, we select the nearest feature statistics according to style-embeddings in the dual-normalization module to normalize target domain features for generalization. Our method shows significant improvement compared to other state-of-the-art methods on BraTS, Cardiac and Abdominal Multi-Organ datasets.

\clearpage

{\small
\bibliographystyle{ieee_fullname}
\bibliography{egbib}
}

\section{Supplement Material}

\subsection{Efficacy of Style Augmentation on Cardiac and Multi-Organ Datasets}
\begin{figure}[h]
    \centering
    \begin{subfigure}{0.49\linewidth}
        \centering
        \includegraphics[width=1.0\columnwidth]{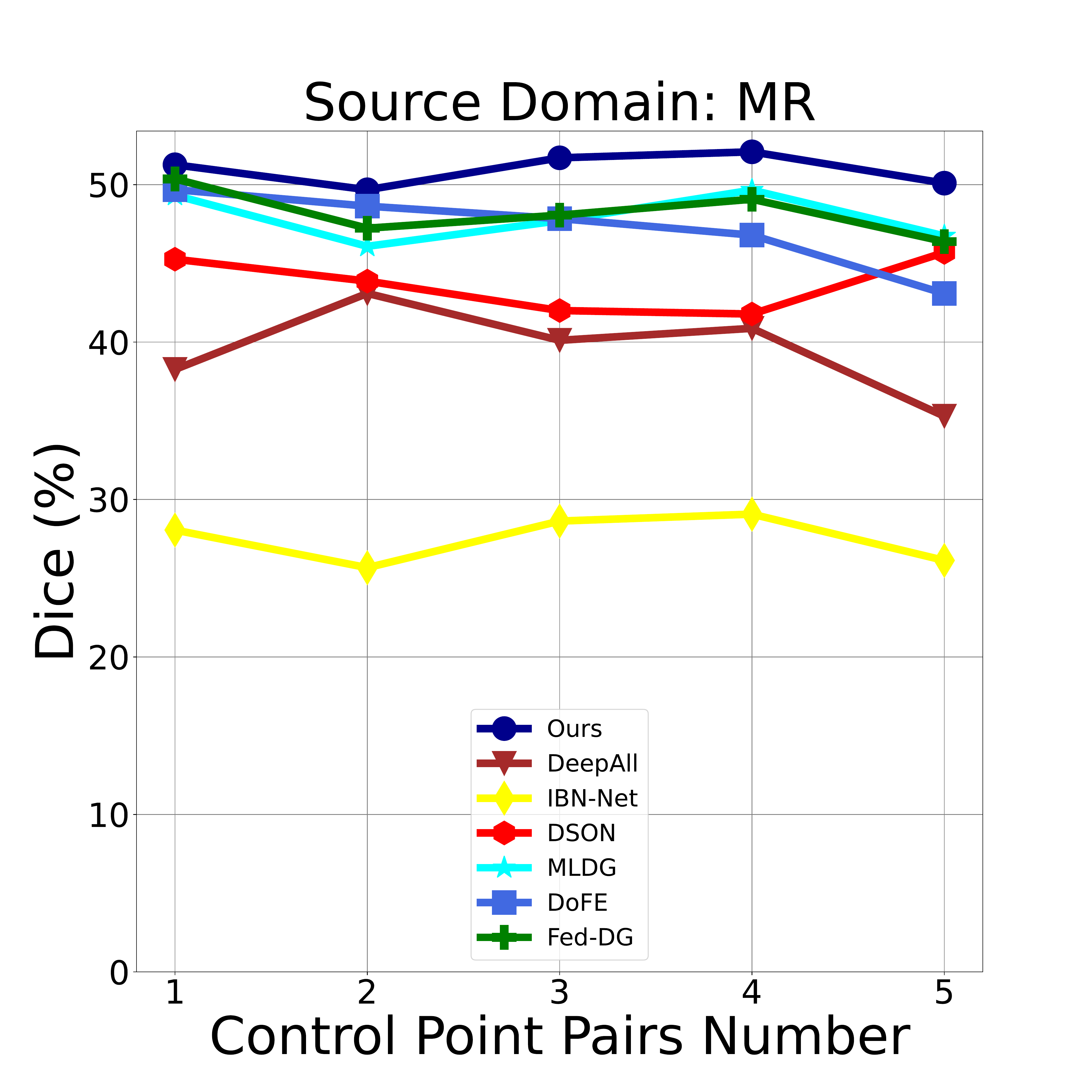}
        \caption{Source Domain T2}
        \label{fig9_a}
    \end{subfigure}
    \begin{subfigure}{0.49\linewidth}
        \centering
        \includegraphics[width=1.0\columnwidth]{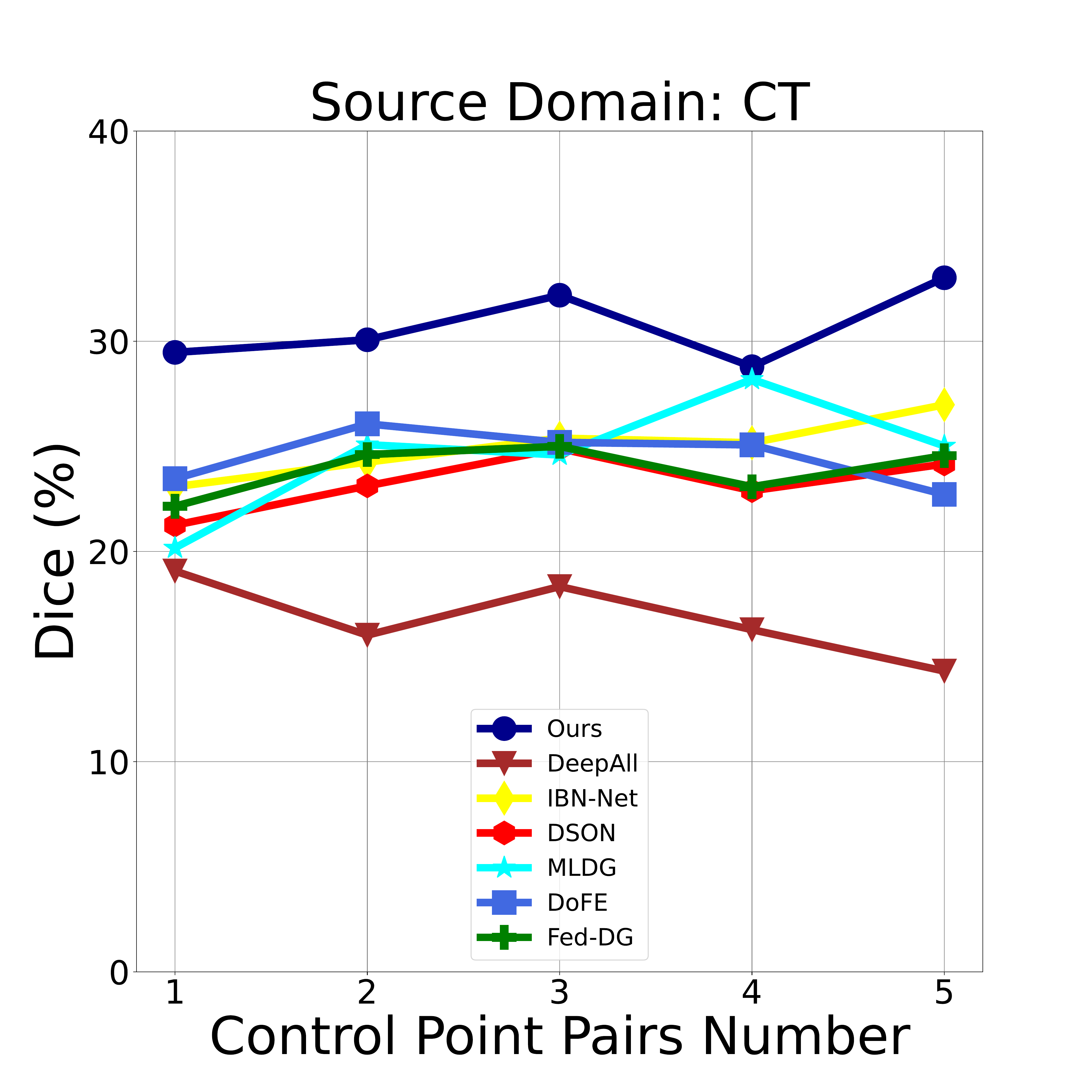}
        \caption{Source Domain T1CE}
        \label{fig9_b}
    \end{subfigure}
    \caption{The segmentation performance on Cardiac dataset of our method and other SOTA methods based on different numbers of control point pairs.}
    \label{cardiac}
    \vspace{-15pt}
\end{figure}

\begin{figure}[h]
    \centering
    \begin{subfigure}{0.49\linewidth}
        \centering
        \includegraphics[width=1.0\columnwidth]{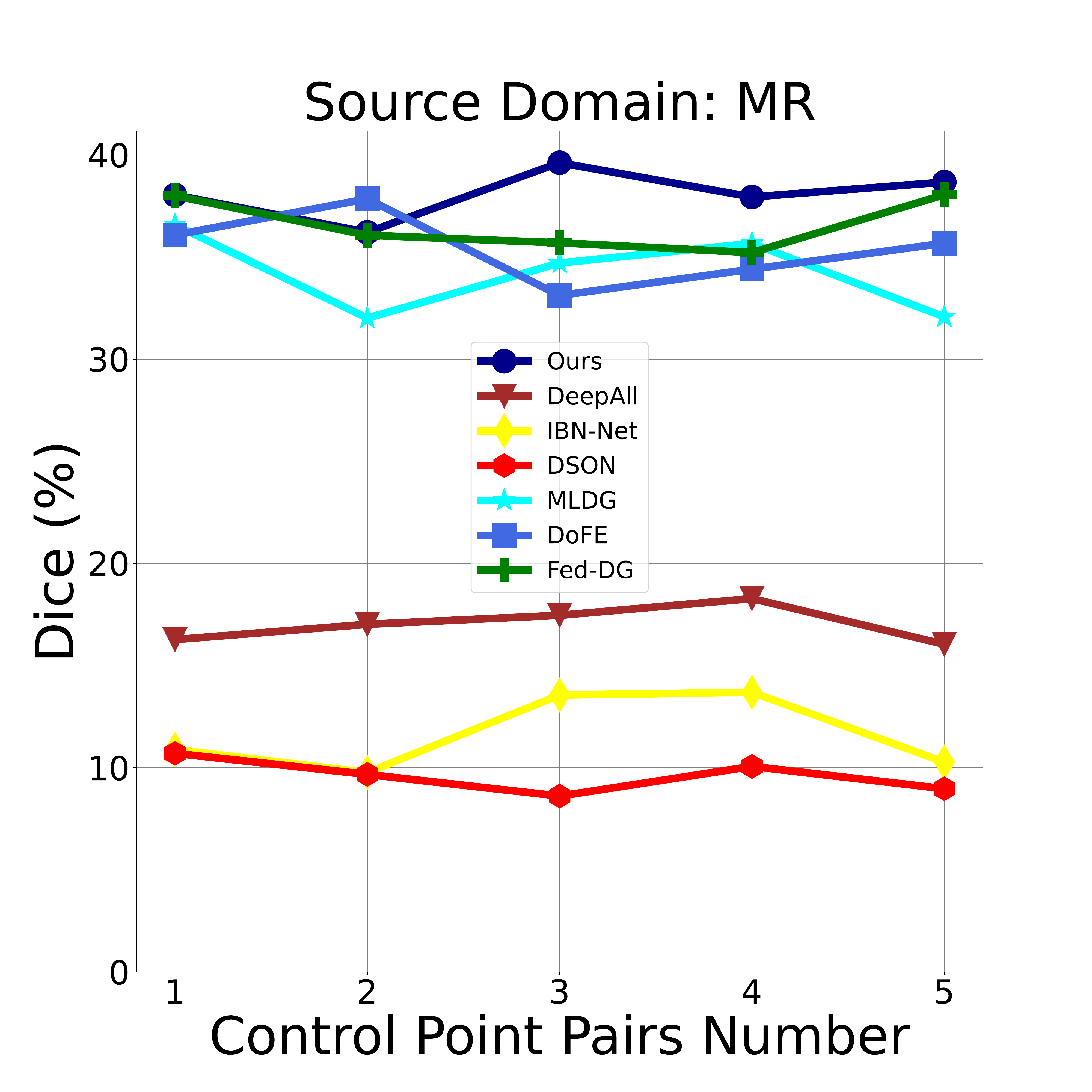}
        \caption{Source Domain T2}
        \label{fig10_a}
    \end{subfigure}
    \begin{subfigure}{0.49\linewidth}
        \centering
        \includegraphics[width=1.0\columnwidth]{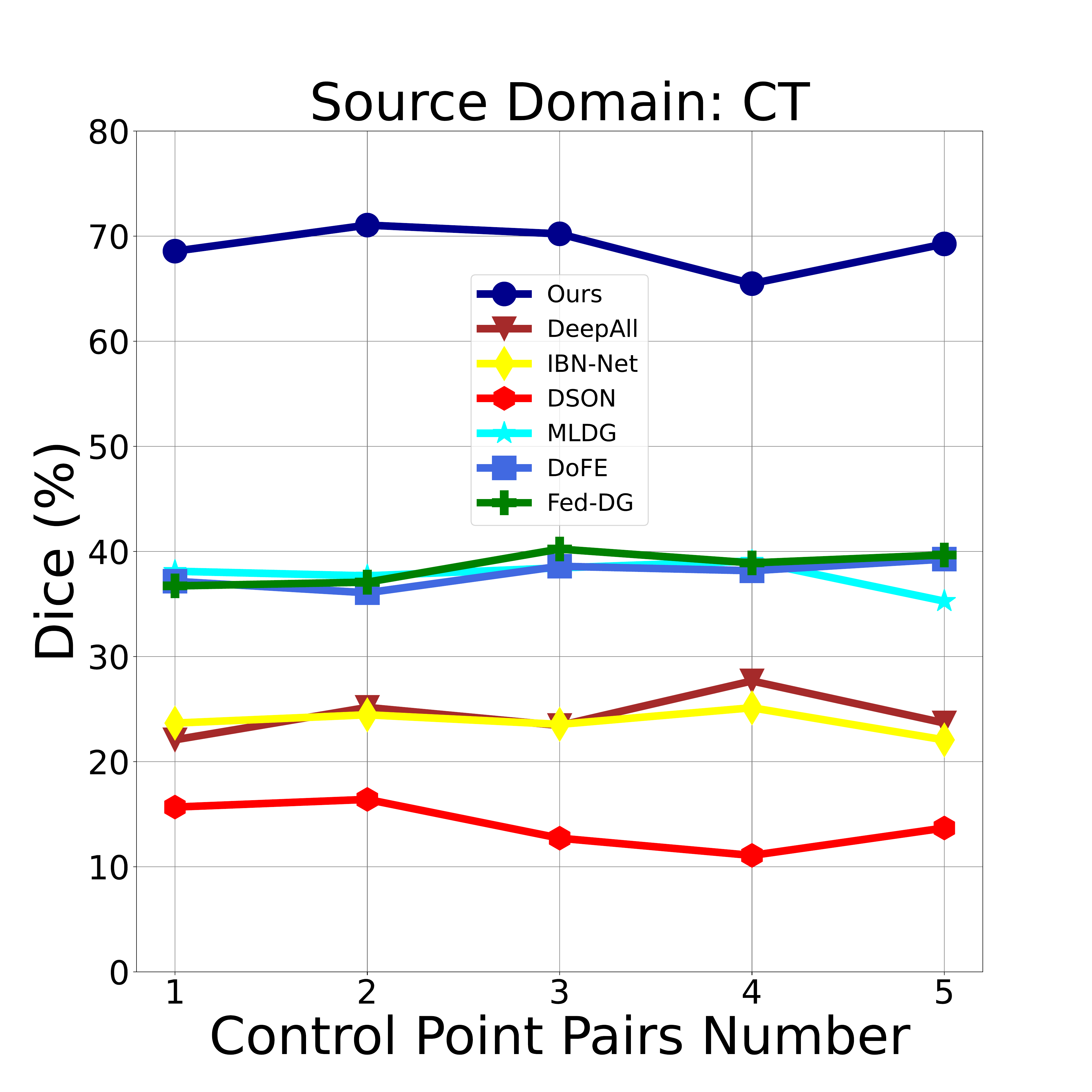}
        \caption{Source Domain T1CE}
        \label{fig10_b}
    \end{subfigure}
    \caption{The segmentation performance on Abdominal Multi-Organ dataset of our method and other SOTA methods based on different numbers of control point pairs.}
    \label{multi-organ}
\end{figure}

We conduct ablation study on the number of control point pairs of B\'ezier Curve on Cardiac and Abdominal Multi-Organ datasets. Similar to the analysis in Section 4.3, we explore different numbers of control point pairs from 1 to 5 in Figure~\ref{cardiac} and \ref{multi-organ}. We can observe that, in most cases, our method can outperform other methods and the results of our method do not fluctuate greatly due to the change of the number of control point pairs. This can further prove that the number of transformation functions will not affect the results of our method a lot if we control the number in a proper range.

\subsection{Visualization of Results on BraTS Dataset}

\begin{figure*}[h]
\centering
\includegraphics[width=1.8\columnwidth]{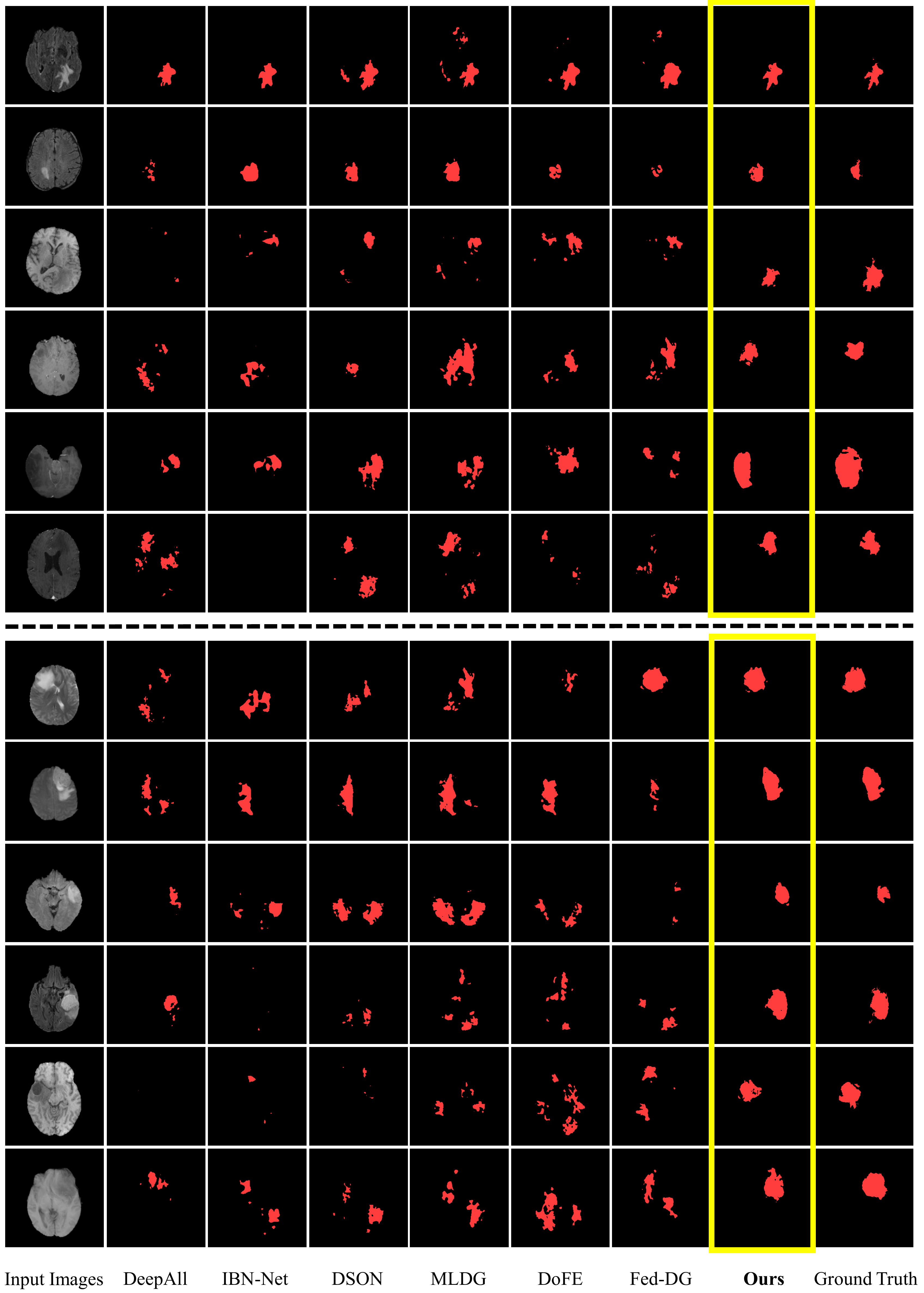}
\caption{Visualization results on BraTS dataset of different methods. First six rows use T2 as source domain; Last six rows use T1CE as source domain.}
\label{fig11}
\vspace{-15pt}
\end{figure*}

We visualize the segmentation results on BraTS Dataset of our method and other compared methods in Figure~\ref{fig11}. We use yellow boxes to highlight our results. It is evident that the segmentation results of other methods are very terrible. More serious, there is even no overlap at all between segmentation masks of some methods and ground truth masks. In contrast, the segmentation masks of our methods not only have high ratio of overlap with ground truth masks but also have good spatial continuity. Our segmentation results are also more similar to ground truth masks in morphology.

\end{document}